\documentclass[runningheads]{llncs}
\usepackage[T1]{fontenc}
\usepackage{booktabs}
\usepackage{hyperref}
\usepackage{graphicx}
\usepackage{multirow}
\usepackage[table]{xcolor}
\usepackage{glossaries}
\usepackage{subcaption}

\usepackage{makecell}
\begin{document}
\title{Exploring the impact of fairness-aware criteria in AutoML}
%
%
\author{Joana Simões\orcidID{0000-0002-6846-3517} \and
João Correia\orcidID{0000-0001-5562-1996}}
\authorrunning{J. Simões and J. Correia}
%
\institute{University of Coimbra, CISUC/LASI – Centre for Informatics and Systems of the University of Coimbra\\
\email{{\{joanasimoes,jncor\}}@dei.uc.pt}}

\maketitle              

\newcolumntype{C}[1]{>{\centering\arraybackslash}m{#1}}

\newacronym{greenautoml}{Green AutoML}{Green Automated Machine Learning}
\newacronym{automl}{AutoML}{Automated Machine Learning}
\newacronym{dcai}{DCAI}{Data Centric Artificial Intelligence}
\newacronym{greenai}{Green AI}{Green Artificial Intelligence}
\newacronym{ml}{ML}{Machine Learning}
\newacronym{ea}{EA}{Evolutionary Algorithm}
\newacronym{edca}{EDCA}{Evolutionary Data Centric AutoML}
\newacronym{dr}{DR}{Data Reduction}
\newacronym{fs}{FS}{Feature Selection}
\newacronym{is}{IS}{Instance Selection}
\newacronym{hpo}{HPO}{Hyperparamenter Optimisation}
\newacronym{ai}{AI}{Artificial Intelligence}
\newacronym{ga}{GA}{Genetic Algorithm}
\newacronym{ec}{EC}{Evolutionary Computation}
\newacronym{mcc}{MCC}{Matthews Correlation Coefficient}
\newacronym{tpr}{TPR}{True Positive Rate}
\newacronym{dp}{DP}{Demographic Parity}
\newacronym{eo}{EO}{Equalised Odds}
\newacronym{abroca}{ABROCA}{Absolute Between-ROC Area}

\begin{abstract}
\gls{ml} systems are increasingly used to support decision-making processes that affect individuals. However, these systems often rely on biased data, which can lead to unfair outcomes against specific groups. 
With the growing adoption of \gls{automl}, the risk of intensifying discriminatory behaviours increases, as most frameworks primarily focus on model selection to maximise predictive performance. 
Previous research on fairness in \gls{automl} had largely followed this trend, integrating fairness awareness only in the model selection or hyperparameter tuning, while neglecting other critical stages of the \gls{ml} pipeline.
This paper aims to study the impact of integrating fairness directly into the optimisation component of an \gls{automl} framework that constructs complete \gls{ml} pipelines, from data selection and transformations to model selection and tuning. 
As selecting appropriate fairness metrics remains a key challenge, our work incorporates complementary fairness metrics to capture different dimensions of fairness during the optimisation.
Their integration within \gls{automl} resulted in measurable differences compared to a baseline focused solely on predictive performance. 
Despite a 9.4\% decrease in predictive power, the average fairness improved by 14.5\%, accompanied by a 35.7\% reduction in data usage. 
Furthermore, fairness integration produced complete yet simpler final solutions, suggesting that model complexity is not always required to achieve balanced and fair \gls{ml} solutions.

\end{abstract}

\glsresetall

\section{Introduction}
\label{sec:introduction}

The increasing adoption of \gls{ml} systems across diverse domains such as healthcare, finance and education has created a demand for efficient and accessible development of these systems. To address it, \gls{automl} emerged to automate the creation, design, and optimisation of \gls{ml} pipelines, accelerating deployment and reducing experts' effort \cite{Zoller_2021_Benchmark}.
However, most solutions focus mainly on tuning the models and their hyperparameters, overlooking other crucial stages of the \gls{ml} workflow. 

As autonomous decision-making systems are increasingly used to make life-changing decisions regarding humans, such as hiring and lending, the risk of discriminatory outcomes becomes more evident.
These issues often come from biased historical data, causing models to learn and reproduce undesirable patterns that disadvantage specific groups \cite{Suresh_2021_Framework}. 
In response, \gls{ai} fairness research emerged to mitigate discrimination and promote equitable \gls{ml} systems, defining fairness as the absence of any prejudice or favouritism towards an individual or a group based on their characteristics \cite{Mehrabi_2019_Survey}.
However, integrating fairness methods into \gls{ml} remains challenging due to the context-dependent nature of fairness definitions and the limited generalisation capability of the existing metrics \cite{Caton_2024_Fairness}.
Moreover, bias can emerge in multiple parts of the \gls{ml} workflow, each requiring a different mitigation strategy \cite{Suresh_2021_Framework}. 
Since these strategies often demand substantial manual effort and experimentation, integrating them into \gls{automl} is a promising direction towards more balanced and comprehensive \gls{ml} solutions.

Unfairness mitigation is complex and time-consuming, and implementing these strategies is usually a manual and resource-intensive process. Therefore, using \gls{automl} to automate fairness optimisation is a natural next step. 
Yet, previous studies suggest that a fully fair \gls{automl} framework is currently unfeasible, as fairness definitions are vague, context-dependent, and sometimes conflicting \cite{Weerts_2024_Fairness}. They suggest that the direction is fairness-aware \gls{automl} approaches that promote fairness into the optimisation while still applying user input for domain-specific decisions, such as defining sensitive attributes and fairness targets.
The main challenge, therefore, lies in balancing fairness and predictive performance during optimisation to achieve an optimal trade-off between the two.

This paper analyses the influence of a fairness-aware optimisation in an \gls{automl} framework by integrating both pre- and in-processing mitigation techniques \cite{Suresh_2021_Framework} in the \gls{ml} pipeline. 
Previous studies adopted a model-centric perspective, focusing solely on model tuning, often restricting optimisation to a single architecture \cite{Cruz_2021_Promoting,Wuwang_2021_Fairautoml,Perrone_2021_FairBO}. 
Instead, our work optimises the entire \gls{ml} pipeline, from data selection to model optimisation, according to the characteristics of the data and user input. The optimisation uses a multi-criteria single-objective genetic algorithm to manage the trade-offs between predictive performance and fairness metrics to guide the optimisation. 

Our work \footnote{The GitHub repository can be found in \url{https://github.com/PugtgYosuky/EDCA}} focus on understanding the impact of integrating fairness into \gls{automl} by comparing it with a traditional \gls{automl} process intended solely to maximise predictive performance. To incorporate fairness into \gls{automl}, the optimisation integrates three complementary fairness metrics (\gls{dp} \cite{Dwork_2012_Fairness}, \gls{eo} \cite{Hardt_2016_Equality}, \gls{abroca} \cite{Gardner_2019_Evaluating}) along with two predictive performance metrics (\gls{mcc} and \gls{tpr}) into a single objective. Both our Baseline and our Fairness-aware setup create complete pipelines and transform the input data. However, the different evaluation functions guide the solutions towards different directions. 
We evaluated the predictive performance, fairness and data characteristics of the final solutions produced by both the Baseline and Fairness-aware setup using three datasets commonly employed in fairness research. 
The results revealed clear differences not only in the metrics' values achieved but also in the structure of the final \gls{ml} pipelines. 
This analysis of how fairness influences the design of \gls{ml} pipelines contributes to \gls{ml} fairness research by illustrating its tangible impact on the resulting \gls{ml} solutions. Furthermore, this work advances \gls{automl} research by highlighting the importance of optimising the entire \gls{ml} workflow, enabling the creation of more balanced, efficient, and fairer \gls{ml} solutions.

The remainder of the paper is organised as follows. Section \ref{sec:background} discusses previous work on fairness and \gls{automl}.
Section \ref{sec:integrating-fairness} details our proposed solution to integrate fairness in \gls{automl} systems.
Section \ref{sec:experimentation} presents the experimental setup and discusses the results.
Finally, Section \ref{sec:conclusion} summarises the main findings and outlines directions for future work. 

\section{Related Work}
\label{sec:background}

Fairness research focuses on developing algorithms that ensure legal, regulatory and ethical standards in \gls{ml} decisions \cite{Weerts_2024_Fairness}. 
However, there is no universal concept of fairness in \gls{ml} as it varies according to the context, and detecting bias is challenging since it can emerge throughout the \gls{ml} lifecycle \cite{Suresh_2021_Framework}. 
Several fairness metrics analyse different aspects of the data and predictions, though none fully capture all bias dimensions \cite{Mehrabi_2019_Survey}, and they can sometimes be conflicting \cite{Weerts_2024_Fairness}.
Also, when considering predictive metrics, there is an inherent trade-off between predictive power and fairness, as balancing outcomes across sensitive groups often reduces overall predictive power \cite{Caton_2024_Fairness}.

Despite there not being an universal metric to measure fairness in \gls{ml}, the three most predominant metrics a \gls{dp} \cite{Dwork_2012_Fairness}, \gls{eo} \cite{Hardt_2016_Equality}, and \gls{abroca} \cite{Gardner_2019_Evaluating}.
\gls{dp} \cite{Dwork_2012_Fairness}, also known as statistical parity, measures whether the rate of positive predictions is equal across all sensitive attribute groups, i.e., whether the model’s predictions are independent of the sensitive attribute. On the other hand, \gls{eo} analyses if true and false positive rate of the outcomes is the same for each sensitive attribute category \cite{Hardt_2016_Equality} conditional to the true label. \gls{abroca} extended \gls{eo} by analysing the true and false positive rates across the entire Receiver Operating Characteristic (ROC) curve \cite{Gardner_2019_Evaluating}. It measures the difference in performance along the ROC curve between the demographic groups. Originally, Gardner et al. only measured \gls{abroca} between two demographic groups for one sensitive attribute. Later,  Mangal et al. \cite{Mangal_2024_Implementing} proposed an extension to capture intersections of groups. A comprehensive overview of fairness evaluation metrics in machine learning is provided in the survey by Quy et al. \cite{Quy_2022_survey}.

Mitigating bias often requires interventions throughout the entire \gls{ml} pipeline \cite{Suresh_2021_Framework,Caton_2024_Fairness}, divided into three categories. The first, pre-processing, aims to transform the data used on the models and remove their bias to make it fairer, as predictive models are only as good as the data they use \cite{Whang_2021_Data}. Examples include selecting the data that balances protected and unprotected attributes.
The second, in-processing mitigation, incorporates fairness as a constraint or additional objective during the models' learning process. Lastly, post-processing adjusts predictions after training to improve fairness toward sensitive attributes.
In summary, ensuring fairness in \gls{ml} is a complex and context-dependent process. As \gls{ml} systems increasingly influence decisions, addressing fairness is critical to prevent harm and promote trust in automated decision-making applications.

\label{sec:fairness-in-automl}
\gls{automl} accelerates \gls{ml} development and enables non-experts to build models without extensive expertise \cite{Zoller_2021_Benchmark}.
However, most \gls{automl} frameworks prioritise model selection for improving predictive performance and often overlook data characteristics and ethical implications, which may amplify harm.

A typical \gls{automl} process starts when a user provides a task and dataset, and the framework searches for an optimal \gls{ml} model. While this search improves predictive performance, critical stages of the \gls{ml} pipeline, such as data preprocessing and fairness considerations, are often ignored. 

Integrating unfairness mitigation within \gls{automl} remains challenging, since there is no universal definition of fairness \cite{Caton_2024_Fairness}. 
Instead, there are problem-dependent definitions and many metrics capturing various bias dimensions.
However, they often conflict and cannot be simultaneously satisfied, making fairness inherently context-dependent.
\gls{automl} frameworks can also integrate pre-, in-, and post-processing mitigation mechanisms \cite{Suresh_2021_Framework,Caton_2024_Fairness}. 
Although their inclusion increases the search space size and complexity, it enables fairer and more complete solutions.
A common way of integrating fairness is by incorporating fairness metrics into the optimisation. 
Previous works have explored this fairness integration in \gls{automl}. Perrone et al. \cite{Perrone_2021_FairBO} integrated fairness constraints into the Bayesian hyperparameter optimisation for different \gls{ml} models to define feasible fairness regions while minimising performance loss.
Wu et al. extended the approach with a self-adaptive fairness strategy to decide whether or not to apply an in-processing or post-processing mitigation \cite{Wuwang_2021_Fairautoml}. 
In contrast, Schmucker et al. proposed a multi-objective optimisation incorporating three fairness metrics, \gls{dp}, \gls{eo}, and equal opportunity, alongside predictive performance and time efficiency \cite{Schmucker_2020_Multi}. Later, Cruz et al. replaced the fairness metrics with predictive equality and introduced adaptive weighting mechanisms to dynamically adjust the trade-offs \cite{Cruz_2021_Promoting}. However, these approaches have a model-centric perspective, as most \gls{automl} frameworks, ignoring the role of data on fairness and focusing solely on model tuning. 

Despite substantial progress, researchers from both \gls{automl} and fairness fields argue that fairness cannot be addressed solely as an optimisation problem due to its contextual nature \cite{Weerts_2024_Fairness}. Instead, \gls{automl} frameworks should be fairness-aware and incorporate user domain-knowledge to define sensitive attributes and relevant fairness metrics for each domain.
These ongoing challenges indicate that \gls{automl} still requires further development to mitigate unfairness effectively and integrate ethical and data concerns.

Our experimentation aims to analyse the impact of pre-processing (data selection) and in-processing (performance and fairness-guided optimisation) strategies in \gls{automl}. Instead of using only one metric, we optimise three complementary fairness metrics and calculate them across all sensitive attributes rather than focusing on only one. Initially, the user defines dataset and its sensitive attributes, and then the frameworks searches for an appropriate trade-off solution between fairness and predictive performance.

\section{Integrating Fairness in AutoML}
\label{sec:integrating-fairness}

The proposed methods are implemented on top of \gls{edca}, an \gls{automl} framework developed by Simões et al. \cite{Simoes_2025_EDCA}. \gls{edca} combines data-centric and energy-efficient design principles to automatically generate and optimise \gls{ml} pipelines.
In brief, \gls{edca} analyses the input data to define a problem-specific pipeline structure, including preprocessing to handle data inconsistencies, instance and feature selection to reduce the dataset, and predictive modelling steps to optimise. It then employs a genetic algorithm to optimise these configurations according to user-defined objectives. This approach allows for an efficient exploration of the search space, while tailoring solutions to data-specific characteristics, reducing unnecessary computation and improving interpretability. Details about \gls{edca} can be found in our previous work \cite{Simoes_2025_EDCA}.

As \gls{edca} uses a data-centric approach to enhance data quality through data selection and data processing mechanisms, its inherent data quality enhancement works as a pre-processing unfairness mitigation mechanism. Also, the integration of fairness metrics into the fitness function controls all mitigation mechanisms and works as an in-processing mitigation to adjust the entire \gls{ml} structure according to fairness principles. 
In the pre-processing stage, \gls{is} and \gls{fs} components are optimised to identify the most relevant data subsets. This selection, guided by the fairness metrics integrated in the optimisation, does not explicitly enforce class and attribute balance, but can indirectly reduce bias and improve the representation of sensitive groups, leading to fairer training data and better fairness scores.

At the in-processing level, and as previously mentioned, fairness is incorporated into the optimisation process through the multi-criteria fitness function that combines both the fairness and the performance of the solutions. This encourages the algorithm to search for pipelines that achieve a better trade-off between predictive power and fairness, as the objectives can be conflicting.

Because no single metric can capture all fairness dimensions \cite{Quy_2022_survey}, our evaluation incorporates three fairness metrics, \gls{dp} \cite{Dwork_2012_Fairness}, \gls{eo} \cite{Hardt_2016_Equality}, \gls{abroca} \cite{Gardner_2019_Evaluating}, as defined in Equation \ref{eq:fairness-component-ff}.
These metrics capture distinct dimensions, including group-level parity, equality of opportunity, and performance consistency across subgroups. 
All fairness metrics are computed over intersections of all combinations of sensitive attributes, following the approach of Mangal et al. \cite{Mangal_2024_Implementing}, rather than considering a single attribute in isolation. This addresses a limitation of prior work, which typically optimises fairness with respect to only one sensitive attribute at a time \cite{Quy_2022_survey,Perrone_2021_FairBO}. Specifically, we calculated the metrics based on the difference between the maximum and minimum values across all subgroup combinations, with the objective of minimising disparities among protected groups. This intersectional evaluation enables a more comprehensive assessment of fairness across multiple sensitive attributes simultaneously.

\begin{equation}
    \operatorname{Fairness}(i)
        = \frac{
            \operatorname{DP}(i)
            + \operatorname{EO}(i)
            + \operatorname{ABROCA}(i)
        }{3}
    \label{eq:fairness-component-ff}
\end{equation}

\noindent where all metrics refer to individual $i$, a \gls{ml} pipeline; $\operatorname{DP}(i)$ is the Demographic Parity distance; $\operatorname{EO}(i)$ = Equalised Odds distance; $\operatorname{ABROCA}(i)$ = Absolute Between-ROC Area;

The Fairness-aware evaluation function (Equation \ref{eq:fairness-aware-ff}) also incorporates a component for the predictive performance of each solution to guide the search towards regions with both good fairness and good performance.
This performance component combines \gls{mcc} with \gls{tpr} with equal importance, as described in Equation  \ref{eq:baseline-fitness-function}. \gls{mcc} was selected because it provides a robust and balanced evaluation of performance, particularly for imbalanced binary classification tasks \cite{Chicco_2020_Advantages}, while \gls{tpr} promotes correctly identifying the minority positive class. This combination promotes a balanced performance evaluation, ensuring that the positive class prediction remains relevant, since one way of exploiting the integration of fairness would be to be "fair" by classifying everything with the negative class, resulting in ultimately useless predictions and degradation of the results.
A previous empirical study has shown us that combining these two metrics achieved good predictive performance and fairness results in this context, as \gls{mcc} controls the balance of the predictions while \gls{tpr} ensures the positive class is not disregarded. 

\begin{equation}
    \operatorname{Performance}(i)
        = \frac{
            \operatorname{nMCC}(i)
            + \left( 1 - \operatorname{TPR}(i) \right)
        }{2}
    \label{eq:baseline-fitness-function}
\end{equation}
using the normalised and inverted version of \gls{mcc} so that all metrics are comprehended in the same scale, as described in Equation \ref{eq:mcc-normalization}.
\begin{equation}
    \operatorname{nMCC}(i)
        = 1 - \frac{\operatorname{MCC}(i) + 1}{2}
    \label{eq:mcc-normalization}
\end{equation}

In the end, the evaluation function used for all individuals $i$ is a weighted sum of both performance and fairness components, using a $\alpha$ parameter to control the trade-off between objectives, as defined in Equation \ref{eq:fairness-aware-ff}.

\begin{equation}
    \operatorname{Fairness-Aware}(i)
        = \alpha \, \operatorname{Performance}(i)
        + (1 - \alpha)\, \operatorname{Fairness}(i)
    \label{eq:fairness-aware-ff}
\end{equation}

\section{Experimentation}
\label{sec:experimentation}

We compared two setups to analyse the influence of fairness metrics in \gls{automl}. The first setup (Baseline) uses only predictive performance to guide the search, while the second (Fairness-aware) incorporates fairness objectives alongside predictive performance.
The Baseline approach uses the predictive performance component of the Fairness-aware evaluation function as its evaluation function (see Equation \ref{eq:baseline-fitness-function}), focusing in reducing the predictive error while achieving a robust performance and promoting correctly identified minority classes. 

On the other hand, the Fairness-aware approach (Equation \ref{eq:fairness-aware-ff}) uses an weighted sum of both predictive performance (Equation \ref{eq:baseline-fitness-function}) and fairness (Equation \ref{eq:fairness-component-ff}) components, controlled by an $\alpha$ parameter. This parameter balances the contribution of both objectives and allows the combination of multiple objectives into a single scalar value (Equation \ref{eq:fairness-aware-ff}). For the $\alpha$ parameter, preliminary experiments to empirically evaluate several parametrisation settings indicated that $\alpha = 0.8$ provided the best trade-off between predictive performance and fairness.

For both fairness and predictive performance components, we averaged their aggregated metrics (Equations \ref{eq:fairness-component-ff} and \ref{eq:baseline-fitness-function}). All the metrics are normalised and scaled to [0, 1], where 0 represents the optimal value for our minimisation objective of reducing both predictive error and unfairness. Appropriate transformations were applied to convert all metrics to this scale, specially Equation \ref{eq:mcc-normalization} which originally uses a different range of values.

\begin{table}[t]
    \centering
    \caption{Parametrisation used in the evaluation. (a) lists the parameters used to compare the two setups; (b) lists the common parameters used internally by EDCA to perform the optimisation.}
    \begin{minipage}[t]{0.45\textwidth}
        \subcaption{Evaluation Parameters}
        \label{tab:setup-evaluation-parameters}
        \resizebox{\textwidth}{!}{
    \begin{tabular}{r|r}
    \toprule
    \rowcolor{gray!30}
        \textbf{Parameter} & \textbf{Value} \\
    \midrule
         External Data Division & Cross-validation \\
         K-fold & 5 \\
         Independent Runs & 30 \\
         Stop Criterion & Time Budget \\
         Time budget (minutes) & 30 min \\
         Performance Metric & \gls{mcc}, \gls{tpr} \\
         Fairness Metrics & \gls{dp} \cite{Dwork_2012_Fairness}, \gls{eo} \cite{Hardt_2016_Equality}, \gls{abroca} \cite{Gardner_2019_Evaluating,Mangal_2024_Implementing} \\
         Fitness Function $\alpha$ & 0.8 \\
    \bottomrule
    \end{tabular}
    }
    \end{minipage}
    \begin{minipage}[t]{0.45\textwidth}
    \subcaption{EDCA parameters}
    \label{tab:framework-parameters}
    \resizebox{\textwidth}{!}{
    \begin{tabular}{r|r}
        \toprule
        \rowcolor{gray!30} \textbf{Parameter} & \textbf{Value} \\
        \midrule
        \multicolumn{2}{c}{Process} \\
        \midrule
            Parallel Jobs & 5 \\
            Data Division & Stratified Train-Validation \\
            Validation Percentage  & 25\% \\
        \midrule
        \multicolumn{2}{c}{Optimisation Algorithm} \\
        \midrule
            Probability of Mutation & 0.3\\
             Probability of Crossover & 0.7\\
             Elitism Size & 1\\
             Population Size & 25 \\
             Tournament Size & 3 \\
             Patience (no. generations) & 5 \\
             Percentage of change & 10\% \\
        \bottomrule
            
    \end{tabular}

    }
    \end{minipage}
\end{table}

Except for the fitness function difference, the experiments were run under identical conditions, including datasets, data splitting strategy, time budget for the optimisation, and search space of data transformations and classification models for the frameworks. 
For each run, we divided the dataset using a stratified 5-fold cross-validation. Then, for each division, we provided the training data to the \gls{automl} framework, which used the available time budget to identify an optimal solution. 
After optimisation, the best solution was evaluated on the test set to assess its generalisation capability. This process was repeated across all datasets in 30 independent runs with 30 different random seeds to ensure statistical robustness. 
The 5-fold division across the 30 independent runs resulted in 150 different solutions for each dataset.

To verify the statistical significance of the results obtained for each dataset, the t-test was employed in parametric cases, while the Wilcoxon signed-rank statistical was applied for non-parametric cases. 
A significance level of 95\% was used in the statistical tests, using a Bonferroni correction to reduce Type I errors due to analysing multiple metrics. Table \ref{tab:setup-evaluation-parameters} summarises the evaluation parametrisation, and Table \ref{tab:framework-parameters} details the framework's internal configuration.

We conducted our experiments in three binary classification datasets commonly employed on fairness tasks \cite{Quy_2022_survey}. Table \ref{tab:datasets-details} summarises their characteristics, sensitive attributes, and respective sensitive group and class proportions. The maximum and minimum proportions refer to the prevalence of sensitive groups in the positive class. 
For the numerical "age" attribute, the framework discretised values into three categories: "<25 years", "25-60 years", ">60 years", similar to Quy et al. \cite{Quy_2022_survey}. \gls{dp} and \gls{eo} fairness metrics were calculated using the \textit{Fairlearn} framework \footnote{\url{https://fairlearn.org}}, while \gls{abroca} was implemented based on Mangal et al. work \cite{Mangal_2024_Implementing}.

\begin{table}[t!]
    \centering
    \caption{Detailed description of the selected binary classification datasets. "Max Sensitive Prop." and "Min Sensitive Prop." indicate the proportion of the most represented and less represented category of any sensitive attribute, respectively.}
    \resizebox{\textwidth}{!}{
    \begin{tabular}{r|C{3cm}|C{3cm}|C{6cm}}
    \toprule
     \rowcolor{gray!30} \textbf{Attribute} & \textbf{adult} & \textbf{credit-card}  & \textbf{portuguese-bank-marketing (B.M.)} \\
    \midrule
    \#Features & 15 & 24 &  17 \\
    \#Instances & 48842 & 30000 &  45211 \\
    \#Size & 732630 & 720000 &  768587 \\
    Missing values & True & False &  False \\
    Negative Class Proportion & 0.761 & 0.779 & 0.883 \\
    Positive Class Proportion & 0.239 & 0.221 & 0.117 \\
    Protected Features & ['age', 'race', 'sex'] & ['x2', 'x4', 'x3'] &   ['age', 'marital'] \\
    Max Sensitive Prop. & 0.220 & 0.125 &  0.101 \\
    Min Sensitive Prop. & 0.001 & 0.000 &  0.005 \\
    \bottomrule
    \end{tabular}
    }
    \label{tab:datasets-details}
\end{table}

\subsection{Experimental Results}
\label{sec:experimentation-results}

Comparing the Baseline with the Fairness-aware scenario, we observed an average improvement of 14.53\% in fairness (two of the three fairness metrics (\gls{dp} and \gls{eo}) consistently improved, while \gls{abroca} exhibited mixed behaviour) with a corresponding loss of 9.41\% in predictive power (Table \ref{tab:avg-results-metrics}). These results were calculated by averaging the percentages of change (\% Change in Table \ref{tab:avg-results-metrics}) across all fairness and performance metrics.
Additionally, the fairness improvement was accompanied by a substantial reduction in the selected data, with an average decrease of 35.69\% of the overall data, which contributed to fairness gains. This consisted of an average reduction of 34.63\% for instances and a 14.81\% for features. 
These results indicate that it is not necessary to have all the available data to train effective and fair \gls{ml} models, and that selecting the data can help balance the dataset and enhance fairness.
However, this reduction in data may have limited the diversity of examples, potentially affecting the models' generalisation ability, as shown with the drop in predictive performance. 
When considering the best solution per run, the performance gap between the Fairness-aware and Baseline setups is reduced, while comparable or improved fairness levels are still achieved (Tables \ref{tab:avg-results-metrics} and \ref{tab:best-results-metrics}). 

\begin{table}[t!]
    \centering
    \caption{Average (± standard deviation) for the fairness, performance and data-related metrics, with the percentage of change (\% Change) from Baseline to the Fairness-aware setup. Statistically significant differences are shown in bold. "Positive \%" = proportion of the positive class, and "\% Sensitives" = proportion of sensitive attributes from the total of features selected by EDCA. The arrows indicate if each metric indicate should increase ($\uparrow$) or decrease ($\downarrow$). (a) results after an initial average of the results achieved on 5-fold cross-validation per run. (b) only the best solution per run. }
    \label{tab:results}
    \begin{subtable}{\linewidth}
        \caption{When averaging the results across 5-folds in each run.}
        \label{tab:avg-results-metrics}
        \resizebox{\textwidth}{!}{

        \begin{tabular}{c|c|c|c|c|c|c|c|c|c}
        \toprule
        \rowcolor{gray!30} \textbf{Dataset} & \multicolumn{3}{c}{\textbf{adult}} & \multicolumn{3}{c}{\textbf{credit-card}} & \multicolumn{3}{c}{\textbf{portuguese-bank-marketing}} \\
        \rowcolor{gray!30} \textbf{Framework} & \textbf{Baseline} & \textbf{Fairness-Aw.} & \scriptsize \textbf{\% Change} & \textbf{Baseline} & \textbf{Fairness-Aw.} & \scriptsize \textbf{\% Change} & \textbf{Baseline} & \textbf{Fairness-Aw.} & \scriptsize \textbf{\% Change} \\
        \midrule
        \textbf{DP}$\downarrow $& 0.45±0.06 & 0.43±0.08 & \scriptsize -4.21\% & 0.49±0.09 & \textbf{0.34±0.09} & \scriptsize -30.29\% & 0.41±0.03 & \textbf{0.29±0.04} & \scriptsize -28.5\% \\
        \textbf{EO}$\downarrow $& 0.99±0.02 & 0.97±0.04 & \scriptsize -1.37\% & 0.91±0.07 & \textbf{0.68±0.11} & \scriptsize -25.61\% & 0.73±0.07 & \textbf{0.63±0.07} & \scriptsize -13.32\% \\
        \textbf{ABROCA}$\downarrow $& 0.34±0.07 & 0.36±0.06 & \scriptsize +7.55\% & 0.85±0.07 & 0.82±0.09 & \scriptsize -3.83\% & 0.33±0.05 & \textbf{0.34±0.05} & \scriptsize +5.08\% \\
        \midrule
        \textbf{MCC} $\uparrow$ & 0.62±0.00 & \textbf{0.60±0.01} & \scriptsize -3.79\% & 0.40±0.00 & \textbf{0.32±0.04} & \scriptsize -20.37\% & 0.50±0.01 & \textbf{0.44±0.02} & \scriptsize -11.82\% \\
        \textbf{TPR} $\uparrow$ & 0.64±0.01 & \textbf{0.61±0.02} & \scriptsize -4.89\% & 0.37±0.01 & \textbf{0.26±0.04} & \scriptsize -28.64\% & 0.48±0.01 & \textbf{0.39±0.03} & \scriptsize -19.03\% \\
        \textbf{AUCROC}$\uparrow$ & 0.92±0.00 & \textbf{0.91±0.00} & \scriptsize -0.89\% & 0.77±0.00 & \textbf{0.72±0.03} & \scriptsize -6.5\% & 0.93±0.00 & \textbf{0.90±0.02} & \scriptsize -2.63\% \\
        \textbf{F1} $\uparrow$ & 0.70±0.00 & \textbf{0.68±0.02} & \scriptsize -3.36\% & 0.48±0.00 & \textbf{0.36±0.05} & \scriptsize -23.44\% & 0.55±0.01 & \textbf{0.48±0.02} & \scriptsize -12.84\% \\
        \midrule
        \textbf{Instances \%} $\downarrow $& 0.77±0.10 & \textbf{0.52±0.15} & \scriptsize -32.13\% & 0.47±0.16 & 0.47±0.14 & \scriptsize +0.36\% & 0.72±0.13 & \textbf{0.45±0.11} & \scriptsize -37.13\% \\
        \textbf{Features \%} $\downarrow $& 0.96±0.04 & 0.94±0.05 & \scriptsize -2.17\% & 0.81±0.12 & \textbf{0.61±0.13} & \scriptsize -25.45\% & 0.91±0.07 & \textbf{0.75±0.14} & \scriptsize -16.83\% \\
        \textbf{Positive \%} $\uparrow$  & 23.94±0.03 & \textbf{23.83±0.18} & \scriptsize -0.46\% & 22.31±0.24 & \textbf{22.10±0.15} & \scriptsize -0.91\% & 11.73±0.04 & 11.65±0.18 & \scriptsize -0.6\% \\
        \textbf{\% Sensitives} $\downarrow$ & 21.51±1.20 & 21.48±1.51 & \scriptsize -0.13\% & 12.73±2.18 & \textbf{16.75±4.87} & \scriptsize +31.52\% & 11.25±1.66 & \textbf{8.53±3.30} & \scriptsize -24.16\% \\
        \bottomrule
        \end{tabular}
        }
        \vspace{0.8em}
        \begin{subtable}{\linewidth}
        \caption{When selecting the best of each run.}
        \label{tab:best-results-metrics}
        \resizebox{\textwidth}{!}{

        \begin{tabular}{c|c|c|c|c|c|c|c|c|c}
        \toprule
        \rowcolor{gray!30} \textbf{Dataset} & \multicolumn{3}{c}{\textbf{adult}} & \multicolumn{3}{c}{\textbf{credit-card}} & \multicolumn{3}{c}{\textbf{portuguese-bank-marketing}} \\
        \rowcolor{gray!30} \textbf{Framework} & \textbf{Baseline} & \textbf{Fairness-Aw.} & \scriptsize \textbf{\% Change} & \textbf{Baseline} & \textbf{Fairness-Aw.} & \scriptsize \textbf{\% Change} & \textbf{Baseline} & \textbf{Fairness-Aw.} & \scriptsize \textbf{\% Change} \\
        \midrule
        \textbf{DP} $\downarrow$& 0.38±0.03 & 0.36±0.05 & \scriptsize -4.64\% & 0.32±0.08 & \textbf{0.23±0.08} & \scriptsize -30.08\% & 0.36±0.05 & \textbf{0.30±0.08} & \scriptsize -15.67\% \\
        \textbf{EO}$\downarrow$ & 0.97±0.09 & 0.94±0.12 & \scriptsize -2.4\% & 0.72±0.23 & \textbf{0.54±0.24} & \scriptsize -24.93\% & 0.54±0.15 & 0.47±0.12 & \scriptsize -13.11\% \\
        \textbf{ABROCA} $\downarrow$& 0.24±0.06 & 0.25±0.07 & \scriptsize +3.44\% & 0.77±0.20 & 0.71±0.21 & \scriptsize -7.89\% & 0.27±0.06 & 0.28±0.05 & \scriptsize +5.0\% \\
        \midrule
        \textbf{MCC} $\uparrow$ & 0.63±0.01 & \textbf{0.62±0.02} & \scriptsize -2.09\% & 0.41±0.01 & \textbf{0.35±0.04} & \scriptsize -13.12\% & 0.51±0.02 & \textbf{0.47±0.03}  & \scriptsize -7.43\% \\
        \textbf{TPR} $\uparrow$ & 0.65±0.01 & \textbf{0.63±0.02} & \scriptsize -2.56\% & 0.37±0.02 & \textbf{0.30±0.07} & \scriptsize -19.59\% & 0.49±0.02 & \textbf{0.44±0.06} & \scriptsize -11.67\% \\
        \textbf{AUCROC} $\uparrow$& 0.92±0.00 & \textbf{0.92±0.01} & \scriptsize -0.49\% & 0.77±0.01 & \textbf{0.74±0.03} & \scriptsize -4.47\% & 0.93±0.01 & \textbf{0.91±0.02} & \scriptsize -1.62\% \\
        \textbf{F1} $\uparrow$& 0.71±0.01 & \textbf{0.69±0.02} & \scriptsize -1.68\% & 0.48±0.01 & \textbf{0.41±0.06} & \scriptsize -15.22\% & 0.56±0.02 & \textbf{0.51±0.04} & \scriptsize -7.78\% \\
        \midrule
        \textbf{Instances \%} $\downarrow$& 0.76±0.23 & 0.54±0.27 & \scriptsize -28.9\% & 0.47±0.33 & 0.39±0.24 & \scriptsize -16.86\% & 0.72±0.27 & 0.53±0.33 & \scriptsize -25.81\% \\
        \textbf{Features \%} $\downarrow$& 0.96±0.10 & 0.99±0.05 & \scriptsize +2.47\% & 0.80±0.26 & 0.60±0.32 & \scriptsize -25.27\% & 0.89±0.20 & 0.80±0.29 & \scriptsize -9.86\% \\
        \textbf{Positive \%} $\uparrow$& 23.97±0.12 & 23.89±0.28 & \scriptsize -0.33\% & 22.41±0.54 & 22.09±0.46 & \scriptsize -1.43\% & 11.73±0.13 & 11.71±0.25 & \scriptsize -0.22\% \\
        \textbf{\% Sensitives} $\downarrow$& 20.47±3.44 & 21.44±0.90 & \scriptsize +4.74\% & 12.75±5.83 & 17.51±12.29 & \scriptsize +37.3\% & 10.17±4.73 & 8.22±5.94 & \scriptsize -19.2\% \\
        
        \bottomrule
        \end{tabular}
        }
            
        \end{subtable}
    \end{subtable}
\end{table}

\begin{figure}[t]
    \centering
    \includegraphics[width=\linewidth]{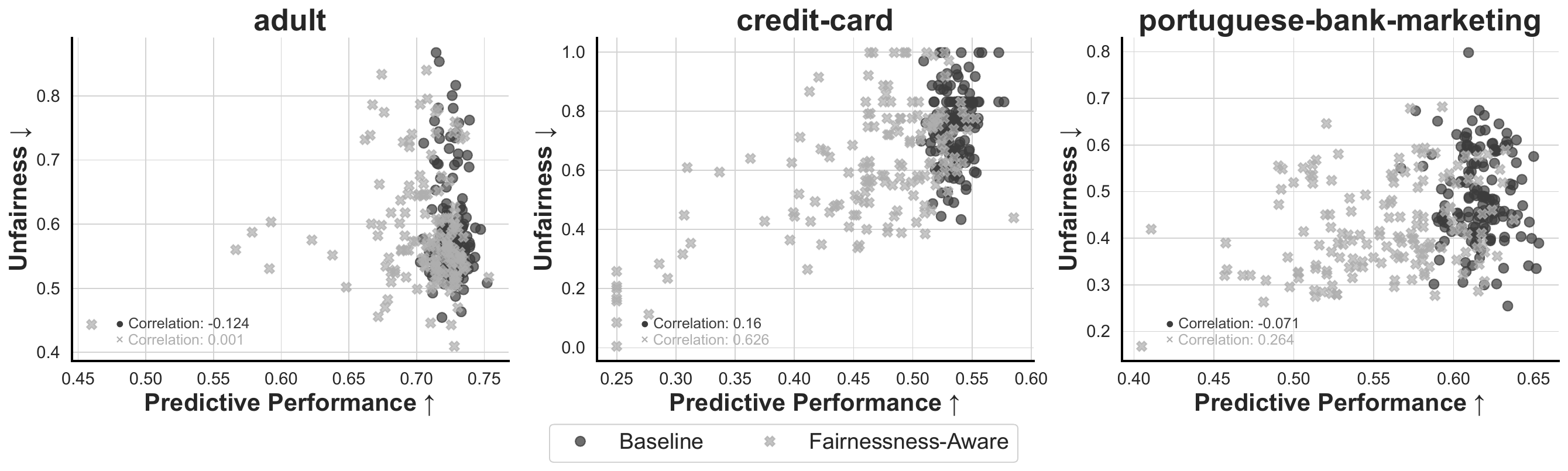}
    \caption{Relationship between fairness (average of \gls{dp} \cite{Dwork_2012_Fairness}, \gls{eo} \cite{Hardt_2016_Equality}, \gls{abroca} \cite{Gardner_2019_Evaluating}) and predictive performance (\gls{mcc} scaled to [0, 1] and \gls{tpr}) measured with the Pearson correlation across both setups for each dataset. Each point represents one of 150 solutions per setup and dataset.}
    \label{fig:fairness-vs-performance}
\end{figure}

Figure \ref{fig:fairness-vs-performance} highlights the trade-off between predictive performance and fairness metrics. The results indicate the metrics achieved in all 150 solutions in each setup for each dataset. The dots show that the Baseline setup achieved an overall better predictive performance, with a concentrated range of values. Nevertheless, despite the increased predictive power, the fairness values are very spread. On the contrary, the Fairness-aware setup (x-shaped markers) showed more disparity of points while being able to achieve better fairness values in some solutions. As expected, the Fairness-aware setup also presented better fairness values with moderate predictive performance. Overall, no strong correlation was observed between fairness and predictive performance, despite the better solutions in terms of fairness always having low predictive power.

\begin{figure}
    \centering
    \includegraphics[width=\textwidth]{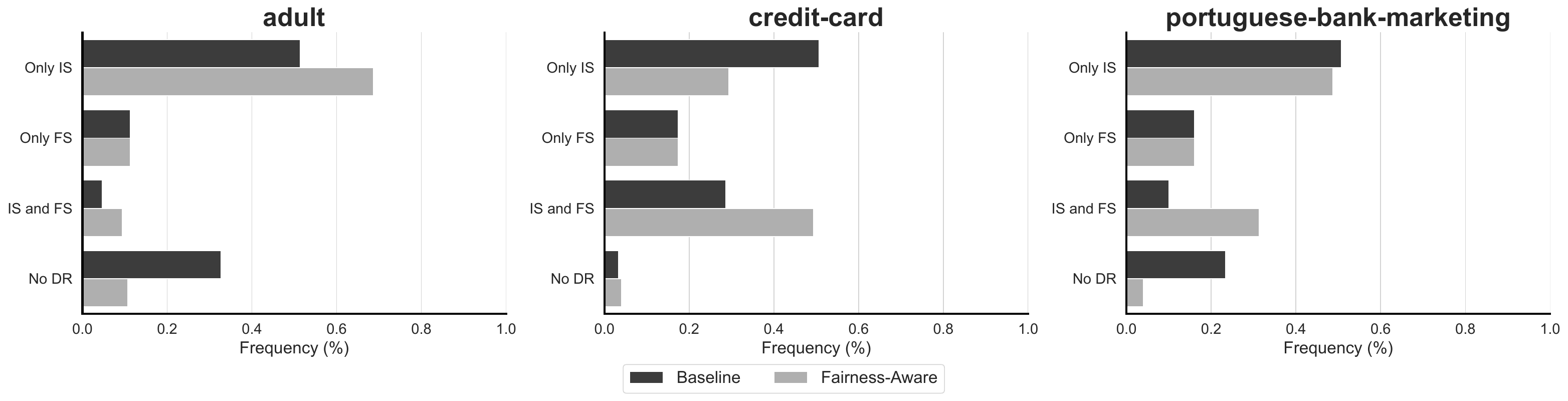}
    \caption{Distribution of the data selection techniques selected on the final solutions across the two setups. It shows how many times each data selection technique appeared on a final solution from the total possible of 150 solutions (5 solutions from the 5-fold cross-validation across 30 seeds).}
    \label{fig:dr-techniques}
\end{figure}

When analysing the data selection techniques used on the end-solutions (Figure \ref{fig:dr-techniques}), we can observe differences across datasets, with clear shifts in the techniques used from the Baseline to the fairness-aware scenario. 
In most cases, the shift reduces the cases without data selection (No DR) and increases the use of instance selection (Only IS) or, particularly, the combination of both instance and feature selection (IS and FS), possibly indicating that rebalancing the data and reducing the features used helps achieve fairness objectives. 
Cases where only \gls{fs} is applied remain identical across both scenarios and all datasets, suggesting that in some situations, altering only the features is sufficient to meet both performance and fairness goals. 
The prevalence of these data selection techniques also explains the substantial reduction in data observed in Table \ref{tab:avg-results-metrics}, as "Only IS" and "IS and FS" are the techniques that reduce the dataset the most. 

\begin{figure}
    \centering
    \includegraphics[width=\textwidth]{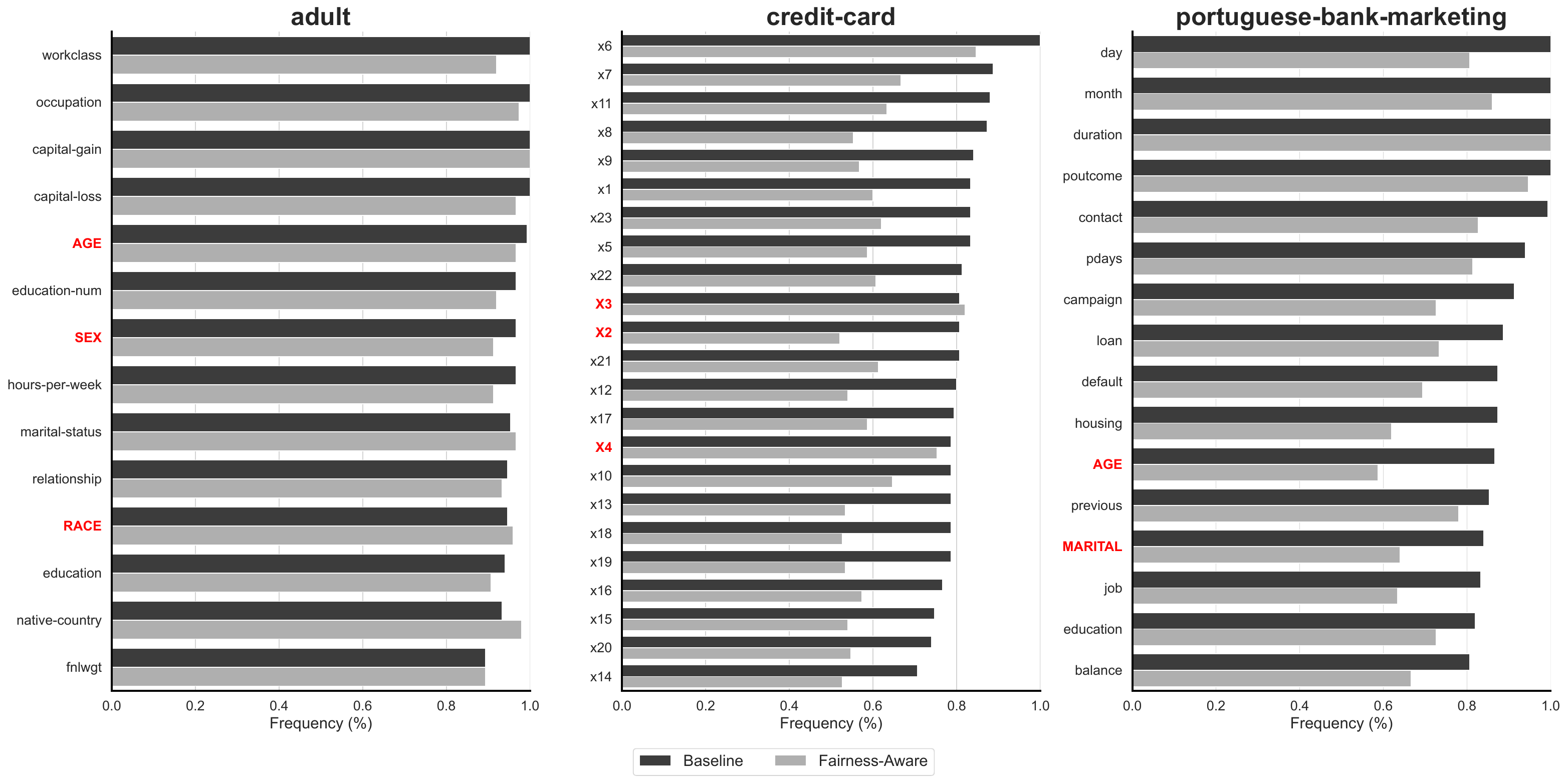}
    \caption{Frequency of each feature on the final solutions. It shows how many times each feature appeared on a final solution from the total possible of 150 solutions (5 solutions from the 5-fold cross-validation across 30 seeds). Highlighted in bold uppercase are the sensitive attributes.}
    \label{fig:feature-importance}
\end{figure}

Despite the substantial reduction in selected instances, the proportions of sensitive attribute categories remained consistent with the original data across both setups, indicating that data selection did not disproportionately affect sensitive groups (Table \ref{tab:sensitive-attributes-prop}). Likewise, the share of sensitive attributes among the selected features shows only minor variation between the Baseline and Fairness-aware setups, with slight reductions in some cases (Tables \ref{tab:avg-results-metrics} and \ref{tab:best-results-metrics}).
At the feature level, the Fairness-aware setup exhibits greater variability in feature selection (Figure \ref{fig:feature-importance}). Sensitive attributes appear less frequently in several datasets, though this behaviour is dataset-dependent, as they remain prevalent in the \textit{credit-card} dataset. Overall, the approach does not enforce the removal of sensitive attributes, but adaptively retains them when they contribute to improved fairness and de-emphasises them otherwise.

\begin{table}[t]
    \centering
    \caption{Proportion of the less represented category of each sensitive attribute. The results indicate the original proportion, and then the average (± standard deviation) proportions for each sensitive attribute category when the attribute was present on the final solutions across all seeds and folds.}
    \resizebox{0.75\textwidth}{!}{
    \begin{tabular}{r|c|c|c|c|c|c}
    \toprule
    \rowcolor{gray!30} \textbf{Dataset} & \textbf{Setup} & \textbf{Age} & \textbf{Race} & \textbf{Gender} & \textbf{Education} & \textbf{Marital} \\
    \midrule
    \multirow{3}{2.5em}{adult} & Original & 0.074 & 0.008 & 0.332 & - & - \\
     & Baseline & 0.074±0.001 & 0.008±0.0 & 0.332±0.002 & - & - \\
     & Fairness-Aware & 0.074±0.002 & 0.008±0.001 & 0.332±0.004 & - & - \\
     \midrule
    \multirow{3}{3em}{credit-card} & Original & - & - & 0.396 & 0.0 & 0.002 \\
     & Baseline & - & - & 0.396±0.006 & 0.001±0.001 & 0.002±0.001 \\
    & Fairness-Aware & - & - & 0.397±0.006 & 0.001±0.001 & 0.002±0.0 \\
    \midrule
    \multirow{3}{6em}{portuguese-bank-marketing} & Original & 0.018 & - & - & - & 0.115 \\
     & Baseline & 0.018±0.001 & - & - & - & 0.115±0.002 \\
     & Fairness-Aware & 0.018±0.001 & - & - & - & 0.115±0.004 \\
    \bottomrule
    \end{tabular}
    }
    \label{tab:sensitive-attributes-prop}
\end{table}

\begin{figure}[t]
    \centering
    \includegraphics[width=\textwidth]{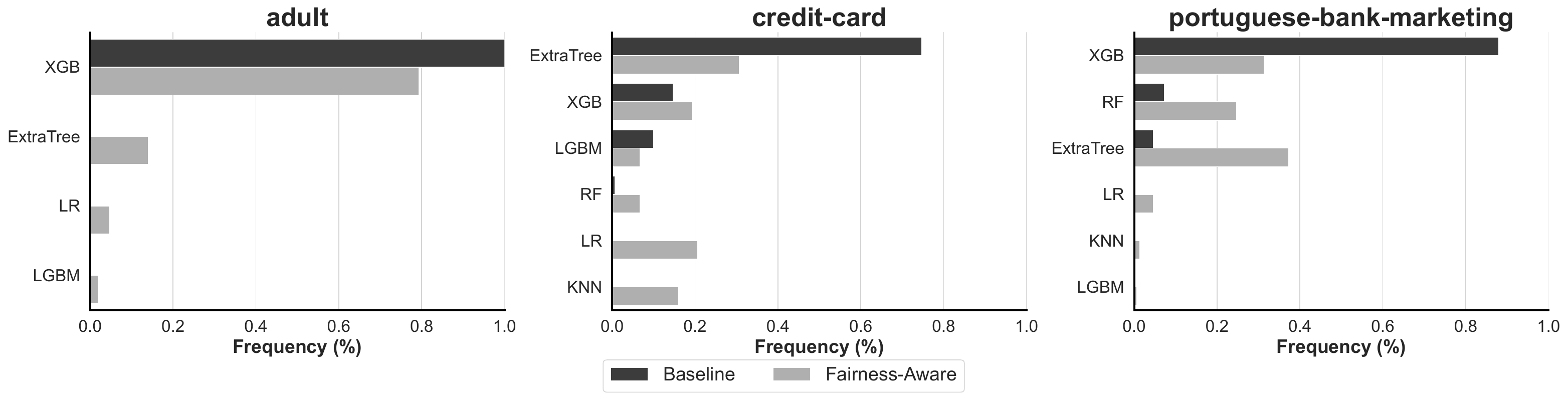}
    \caption{Distribution of the classification models selected on the final solutions across the two setups. It shows how many times each model appeared on a final solution from the total possible of 150 solutions (5 solutions from the 5-fold cross-validation across 30 seeds).}
    \label{fig:classification-models}
\end{figure}

Continuing with the analysis of the final \gls{ml} pipeline solutions, Figure \ref{fig:classification-models} shows clear differences in the models selected. When fairness is introduced, the range of selected models becomes more diverse and shifts towards simpler models, such as logistic regression, compared to the Baseline, which predominantly selected complex ensemble models aimed only to maximise predictive power.

\begin{figure}[t!]
    \centering
    \includegraphics[width=0.75\linewidth]{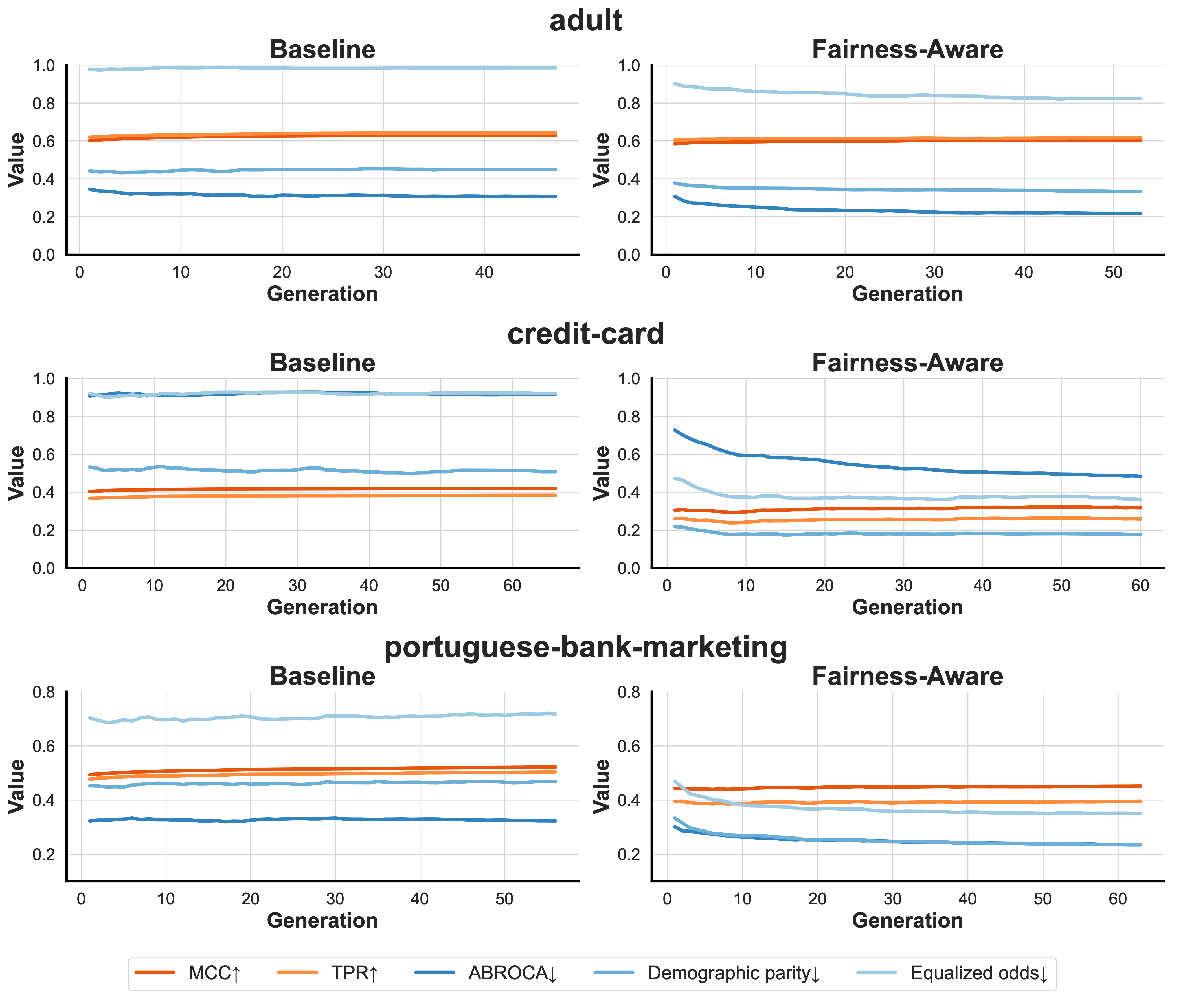}
    \caption{Evolution of the performance and fairness metrics used in the Fitness Function (Equations \ref{eq:baseline-fitness-function} and \ref{eq:fairness-component-ff}). Each metric's line is the average for all the runs and folds, containing the original values without the transformations applied for the fitness function. The graph was cut to the minimum of generations achieved in one of the runs.}
    \label{fig:evolution-metrics}
\end{figure}

Finally, we compared the evolution of both predictive performance and fairness metrics to understand their relationship. Figure \ref{fig:evolution-metrics} shows the average evolution of each metric of the fitness function across all seeds and folds. 
Although the overall trends of \gls{mcc} and \gls{tpr} are similar, the values are consistently shifted, revealing a clear reduction in predictive performance.
In contrast, the fairness metrics behave differently, and improving one metric sometimes worsens another (e.g. relation between \gls{dp} and \gls{eo} in the fairness-aware scenario in the credit-card dataset), highlighting the inherent conflicts between fairness objectives. 
Also, comparing  Baseline and Fairness-aware scenarios reveals clear differences. In the Baseline, despite the increased predictive power, the fairness metrics remain almost unchanged over the generations. In contrast, in the Fairness-aware setup, the fairness metrics improve alongside predictive performance due to the multi-objective evaluation function that integrates both performance and fairness (see Equation \ref{eq:fairness-aware-ff}).

\subsection{Discussion}
\label{sec:experimentation-discussion}

The experimental results confirmed the expected trade-off between predictive performance and fairness in \gls{ml} systems, even when using an \gls{automl} framework (Table \ref{tab:avg-results-metrics}). 
However, introducing fairness metrics into the optimisation led the Fairness-aware setup to achieve an average fairness improvement of 14.53\%, with only a 9.41\% decrease in predictive performance, a similar behaviour to the one found by Cruz et al. work \cite{Cruz_2021_Promoting} and as stated by the literature how the conflicts between both goals \cite{Caton_2024_Fairness}. 
This aligns with previous studies confirming that fairness can slightly reduce predictive power.
Although predictive performance decreased, the magnitude of this reduction was moderate on average, indicating that fairness integration can meaningfully reduce performance disparities across groups while maintaining acceptable predictive performance.
Additionally, the results varied across datasets, indicating that datasets may benefit from different fairness metrics, as proposed by Weerts et al. \cite{Weerts_2024_Fairness} and Quy et al.\cite{Quy_2022_survey}.

Moreover, data selection appeared to have an important impact on fairness, as the Fairness-aware setup has significantly reduced the data used, retaining only a subset of examples and potentially leading to more computationally efficient models. 
Although the data was not explicitly rebalanced, the selection prioritised more informative examples, enhancing efficiency.
Sensitive attributes were often less chosen, and feature diversity increased, suggesting a reduced reliance on sensitive attributes in some datasets, which may help mitigate certain forms of bias, supporting ethical principles (Figure \ref{fig:feature-importance}). Despite being less frequent, their proportion among the features selected increased in one dataset (Table \ref{tab:avg-results-metrics}), indicating that they could be important to some extent both for the fairness and for the predictive performance. Nevertheless, in the other two cases, their usage was reduced in the Fairness-aware setup. This extensive data selection may have reduced the diversity of training examples, which could partially explain the observed decrease in predictive performance.

Another key finding was that by introducing fairness favoured simpler models (Figure \ref{fig:classification-models}), which are often associated with improved generalisation and reduced susceptibility to certain forms of bias. These results reinforce that complex models are not always necessary, and that a fairness-aware optimisation can produce simpler yet effective solutions that are also easier to interpret.

Finally, an analysis of fairness and performance metrics revealed conflicts among fairness metrics, indicating that the different metrics capture distinct fairness dimensions and may contradict each other (Figure \ref{fig:evolution-metrics}). 
Consequently, these findings highlight the importance of employing a multi-objective evaluation that combines multiple predictive performance and fairness metrics, enabling a more comprehensive assessment of the models across multiple perspectives. 

\section{Conclusion}
\label{sec:conclusion}

The widespread use of automated decision-making systems has often intensified discriminatory behaviour, affecting individuals in critical domains. Fairness research emerged to promote ethical \gls{ml} by ensuring bias and model inaccuracies do not disadvantage individuals, especially based on race, gender, or other inherent characteristics. 
As \gls{automl} continues to advance the development of \gls{ml} systems, it must incorporate fairness concerns to avoid amplifying existing biases. 
Although achieving a fully fair \gls{automl} system remains challenging due to the context-dependent nature of fairness definitions, mitigation mechanisms can be embedded within \gls{automl} pipelines supported by user domain knowledge. These can create fairness-aware systems that adapt mitigation strategies while automating optimisation.

This paper evaluated the impact of integrating a fairness-guided search optimisation within an \gls{automl} framework. 
The framework used optimises the entire \gls{ml} pipeline, tailoring its structure to data characteristics and problem requirements. Its data selection component acts as a pre-processing unfairness mitigation mechanism, selecting subsets of relevant instances and features to reduce bias. Furthermore, the integration of fairness into the evaluation function serves as an in-processing mitigation mechanism, adjusting the \gls{ml} pipeline structure throughout the optimisation process. 

Our experiments compared a Fairness-aware setup with a Baseline focused solely on predictive performance across three benchmark datasets.
The results indicate that fairness integration influences not only the overall results achieved, but also the structure of the \gls{ml} solutions, from the data selection mechanisms and data used, to the architecture of the classification models. 
Despite a 9.4\% drop in predictive performance, the Fairness-aware setup achieved a 14.5\% improvement in fairness and reduced data usage by 35.7\%. It also produced simpler yet more complete solutions, highlighting that fairness and accessibility can coexist without excessive model complexity.

Overall, the Fairness-aware \gls{automl} can produce fairer, simpler, and more computationally efficient \gls{ml} solutions through a single-objective evaluation function integrating multiple performance and fairness criteria and data selection mechanisms. However, the results also show the conflicts and trade-offs between fairness, predictive performance and data, indicating that the optimal balance between these objectives may be context-dependent.

Future work should explore integrating post-processing mitigation, analysing its impact on both predictive and fairness metrics and the resulting optimisation complexity.
Further studies should also examine fairness-guided data selection strategies to promote class and sensitive attributes balance, and address possible automation bias \cite{Suresh_2021_Framework} by ensuring that user oversight remains central in fairness-aware \gls{automl} design.

\begin{credits}
\subsubsection{\ackname}
This work is funded by national funds through FCT – Foundation for Science and Technology, I.P., within the scope of the research unit UID/00326 - Centre for Informatics and Systems of the University of Coimbra, 
and through the Portuguese Recovery and Resilience Plan (PRR) through project C645008882-00000055, Center for Responsible AI.
The first author acknowledges FCT under the grant 2025.05405.BD.

\subsubsection{\discintname}
This document was reviewed and refined with the assistance of Large Language Models, such as ChatGPT, or similar tools, which helped check grammar, correct typos, and enhance clarity. 
The overall content and ideas remain solely the responsibility of the authors.
\end{credits}
\bibliographystyle{splncs04}
\bibliography{references}

@inproceedings{Suresh_2021_Framework,
 author = {Harini Suresh and
John V. Guttag},
 bibsource = {dblp computer science bibliography, https://dblp.org},
 booktitle = {EAAMO 2021: ACM Conference on Equity and Access in Algorithms,
Mechanisms, and Optimization, Virtual Event, USA, October 5 - 9, 2021},
 doi = {10.1145/3465416.3483305},
 pages = {17:1--17:9},
 publisher = {ACM},
 title = {A Framework for Understanding Sources of Harm throughout the Machine
Learning Life Cycle},
 year = {2021}
}

@article{Mehrabi_2019_Survey,
author       = {Ninareh Mehrabi and
                  Fred Morstatter and
                  Nripsuta Saxena and
                  Kristina Lerman and
                  Aram Galstyan},
  title        = {A Survey on Bias and Fairness in Machine Learning},
  journal      = {{ACM} Comput. Surv.},
  volume       = {54},
  number       = {6},
  pages        = {115:1--115:35},
  year         = {2022},
  doi          = {10.1145/3457607},
  bibsource    = {dblp computer science bibliography, https://dblp.org}
}

@article{Quy_2022_survey,
 author = {Tai Le Quy and
Arjun Roy and
Vasileios Iosifidis and
Wenbin Zhang and
Eirini Ntoutsi},
 bibsource = {dblp computer science bibliography, https://dblp.org},
 doi = {10.1002/WIDM.1452},
 journal = {WIREs Data Mining Knowl. Discov.},
 number = {3},
 title = {A survey on datasets for fairness-aware machine learning},
 volume = {12},
 year = {2022}
}

@article{Weerts_2024_Fairness,
 author = {Hilde J. P. Weerts and
Florian Pfisterer and
Matthias Feurer and
Katharina Eggensperger and
Edward Bergman and
Noor H. Awad and
Joaquin Vanschoren and
Mykola Pechenizkiy and
Bernd Bischl and
Frank Hutter},
 bibsource = {dblp computer science bibliography, https://dblp.org},
 doi = {10.1613/JAIR.1.14747},
 journal = {J. Artif. Intell. Res.},
 pages = {639--677},
 title = {Can Fairness be Automated? Guidelines and Opportunities for Fairness-aware
AutoML},
 volume = {79},
 year = {2024}
}

@article{Whang_2021_Data,
  author       = {Steven Euijong Whang and
                  Yuji Roh and
                  Hwanjun Song and
                  Jae{-}Gil Lee},
  title        = {Data collection and quality challenges in deep learning: a data-centric
                  {AI} perspective},
  journal      = {{VLDB} J.},
  volume       = {32},
  number       = {4},
  pages        = {791--813},
  year         = {2023},
  doi          = {10.1007/S00778-022-00775-9},
  bibsource    = {dblp computer science bibliography, https://dblp.org}
}

@inproceedings{Gardner_2019_Evaluating,
 author = {Josh Gardner and
Christopher Brooks and
Ryan Baker},
 bibsource = {dblp computer science bibliography, https://dblp.org},
 booktitle = {Proceedings of the 9th International Conference on Learning Analytics
and Knowledge, LAK 2019, Tempe, AZ, USA, March 4-8, 2019},
 doi = {10.1145/3303772.3303791},
 pages = {225--234},
 publisher = {ACM},
 title = {Evaluating the Fairness of Predictive Student Models Through Slicing
Analysis},
 year = {2019}
}

@article{Mangal_2024_Implementing,
 author = {Mudit Mangal and
Zachary A. Pardos},
 bibsource = {dblp computer science bibliography, https://dblp.org},
 doi = {10.1111/BJET.13484},
 journal = {Br. J. Educ. Technol.},
 number = {5},
 pages = {2003--2038},
 title = {Implementing equitable and intersectionality-aware ML in education:
A practical guide},
 volume = {55},
 year = {2024}
}

@inproceedings{Dwork_2012_Fairness,
  author       = {Cynthia Dwork and
                  Moritz Hardt and
                  Toniann Pitassi and
                  Omer Reingold and
                  Richard S. Zemel},
  editor       = {Shafi Goldwasser},
  title        = {Fairness through awareness},
  booktitle    = {Innovations in Theoretical Computer Science 2012, Cambridge, MA, USA,
                  January 8-10, 2012},
  pages        = {214--226},
  publisher    = {{ACM}},
  year         = {2012},
  doi          = {10.1145/2090236.2090255},
  bibsource    = {dblp computer science bibliography, https://dblp.org}
}

@article{Hardt_2016_Equality,
  title={Equality of opportunity in supervised learning},
  author={Hardt, Moritz and Price, Eric and Srebro, Nati},
  journal={Advances in neural information processing systems},
  volume={29},
  year={2016}
}

@article{Caton_2024_Fairness,
 author = {Simon Caton and
Christian Haas},
 bibsource = {dblp computer science bibliography, https://dblp.org},
 doi = {10.1145/3616865},
 journal = {ACM Comput. Surv.},
 number = {7},
 pages = {166:1--166:38},
 title = {Fairness in Machine Learning: A Survey},
 volume = {56},
 year = {2024}
}

@article{Zoller_2021_Benchmark,
 author = {Marc-André Zöller and
Marco F. Huber},
 bibsource = {dblp computer science bibliography, https://dblp.org},
 doi = {10.1613/JAIR.1.11854},
 journal = {J. Artif. Intell. Res.},
 pages = {409--472},
 title = {Benchmark and Survey of Automated Machine Learning Frameworks},
 volume = {70},
 year = {2021}
}

@inproceedings{Simoes_2025_EDCA,
  author       = {Joana Sim{\~{o}}es and
                  Jo{\~{a}}o Correia},
  editor       = {Pablo Garc{\'{\i}}a{-}S{\'{a}}nchez and
                  Emma Hart and
                  Sarah L. Thomson},
  title        = {{EDCA} - An Evolutionary Data-Centric AutoML Framework for Efficient
                  Pipelines},
  booktitle    = {Applications of Evolutionary Computation - 28th European Conference,
                  EvoApplications 2025, Held as Part of EvoStar 2025, Trieste, Italy,
                  April 23-25, 2025, Proceedings, Part {II}},
  series       = {Lecture Notes in Computer Science},
  volume       = {15613},
  pages        = {71--88},
  publisher    = {Springer},
  year         = {2025},
  doi          = {10.1007/978-3-031-90065-5\_5},
  bibsource    = {dblp computer science bibliography, https://dblp.org}
}

@inproceedings{Wuwang_2021_Fairautoml,
    title={Fair AutoML},
    author={Qingyun Wu and Chi Wang},
    year={2021},
    booktitle={ArXiv preprint arXiv:2111.06495},
}

@inproceedings{Perrone_2021_FairBO,
author = {Perrone, Valerio and Donini, Michele and Zafar, Muhammad Bilal and Schmucker, Robin and Kenthapadi, Krishnaram and Archambeau, C\'{e}dric},
title = {Fair Bayesian Optimization},
year = {2021},
isbn = {9781450384735},
publisher = {Association for Computing Machinery},
address = {New York, NY, USA},
doi = {10.1145/3461702.3462629},
booktitle = {Proceedings of the 2021 AAAI/ACM Conference on AI, Ethics, and Society},
pages = {854–863},
numpages = {10},
location = {Virtual Event, USA},
series = {AIES '21}
}

@article{Schmucker_2020_Multi,
  title={Multi-objective multi-fidelity hyperparameter optimization with application to fairness},
  author={Schmucker, Robin and Donini, Michele and Perrone, Valerio and Archambeau, C{\'e}dric},
  year={2020}
}

@inproceedings{Cruz_2021_Promoting,
  title={Promoting fairness through hyperparameter optimization},
  author={Cruz, Andr{\'e} F and Saleiro, Pedro and Bel{\'e}m, Catarina and Soares, Carlos and Bizarro, Pedro},
  booktitle={2021 IEEE international conference on data mining (ICDM)},
  pages={1036--1041},
  year={2021},
  organization={IEEE}
}

@article{Chicco_2020_Advantages,
  title={The advantages of the Matthews correlation coefficient (MCC) over F1 score and accuracy in binary classification evaluation},
  author={Chicco, Davide and Jurman, Giuseppe},
  journal={BMC genomics},
  volume={21},
  number={1},
  pages={6},
  year={2020},
  publisher={Springer}
}
\end{document}